\documentclass[11pt]{article}
\usepackage[final]{acl}
\usepackage{times}
\usepackage{latexsym}
\usepackage[T1]{fontenc}
\usepackage[utf8]{inputenc}
\usepackage{microtype}
\usepackage{inconsolata}

\usepackage{graphicx}
\usepackage{lipsum}
\usepackage{amssymb}
\usepackage{amsmath}
\usepackage{tcolorbox}
\usepackage{multirow}
\usepackage{booktabs}
\usepackage{xcolor}
\usepackage{colortbl}
\usepackage{array}
\usepackage{subcaption}
\usepackage{amssymb}
\definecolor{selected}{RGB}{255, 243, 205}
\definecolor{colA}{RGB}{192, 0, 0}     
\definecolor{colB}{RGB}{46, 117, 182}   
\definecolor{colC}{RGB}{56, 142, 60}  
\definecolor{oursgray}{RGB}{220,220,220}

\title{Improving Lexical Difficulty Prediction with Context-Aligned Contrastive Learning and Ridge Ensembling}
\author{
    Wicaksono Leksono Muhamad$^{\dagger,1}$,
    Joanito Agili Lopo$^{\dagger,1}$,
    Tsamarah Rana Nugraha$^{\heartsuit,1,2}$, \\ 
    \textbf{
    Ahmad Cahyono Adi$^{\heartsuit,1,3}$,
    Muhammad Oriza Nurfajri$^{3}$
    }\\
    $^1$Mantera Studio \quad
    $^2$The University of Manchester \quad
    $^3$Universitas Gadjah Mada \\
    \texttt{\{wcksnlxn,amalopo99,ahmadseverine83\}@gmail.com} \\ 
    \texttt{tsamarah.nugraha@postgrad.manchester.ac.uk} \\
    \texttt{oriza\_nurfajri@mail.ugm.ac.id}
}

\begin{document}
\maketitle
\begingroup
\renewcommand{\thefootnote}{}
\footnotetext{$^\dagger$ Main contributors.}
\footnotetext{$^\heartsuit$ Major contributors.}
\endgroup

\begin{abstract}
Lexical difficulty prediction is a fundamental problem in language learning and readability assessment, requiring models to estimate word difficulty across different first-language (L1) backgrounds. However, existing approaches rely on regression-only training with scalar supervision, which does not explicitly structure the representation space, limiting their ability to capture cross-lingual alignment and ordinal difficulty. To mitigate these issues, we propose \textit{Context-Aligned Contrastive Regression}, which integrates Ridge regression ensemble with two complementary objectives, i.e., Cross-View Context and Ordinal Soft Contrastive Learning. Experiments on three L1 datasets show that (i) contrastive objectives improve cross-lingual representation alignment while preserving language-specific nuances, (ii) the learned representations capture the ordinal structure of lexical difficulty, and (iii) the ensemble effectively mitigates systematic biases of individual models, leading to more stable performance across difficulty levels.\footnote{ 
\url{https://github.com/airlanggawicaksono/BEA2026TOEBM}}
\end{abstract}

\section{Introduction}

Vocabulary is a central component of English as a Foreign Language proficiency, supporting the development of reading, listening, writing, and speaking skills \cite{alshumrani2024unveiling}. Since learners’ proficiency levels influence how vocabulary is acquired \cite{bao2024effects}, estimating lexical difficulty is an important step in the development of level-appropriate learning materials and valid assessment instruments \cite{goyibova2025tailoring}

Prior work has explored related tasks such as Complex Word Identification, Lexical Difficulty Prediction, and Lexical Simplification \cite{paetzold-specia-2016-semeval,yimam-etal-2018-report,shardlow-etal-2021-semeval}. However, they were not tailored for English language learners and ignored how learners' first language (L1) can make English vocabulary easier or harder. This limitation is critical, as vocabulary knowledge is inherently multi-layered and involves several interrelated components that must be acquired for effective language use~\cite{Schmitt2010ResearchingVocabulary}. 

Among the interrelated components, L1 interference plays a central role across linguistic levels, including grammar, syntax, phonology, and vocabulary, leading to systematic differences in how learners from different L1 backgrounds perceive and process words \cite{Alisoy_2024}. Consequently, lexical difficulty is not an intrinsic property of a word, but a relational phenomenon that varies across L1 backgrounds, motivating the need for L1-aware modeling \cite{skidmore-etal-2025-transformer}.

However, learner-specific factors alone are insufficient. The ability to distinguish similar-sounding words does not guarantee successful acquisition or comprehension \cite{pajak2016difficulty}. Therefore, lexical difficulty cannot be reliably inferred from form-level properties alone. Instead, difficulty is shaped by how words are encountered and interpreted in context. Psycholinguistic evidence shows that contextual cues guide meaning interpretation and comprehension \cite{garten2019measuring}. Effective lexical difficulty modeling should therefore capture both learner-specific and contextual dimensions.

To address these challenges, we propose \textit{Context-Aligned Contrastive Regression}, a multi-objective framework for L1-aware lexical difficulty prediction. Our approach (i) integrates direct regression with Cross-View Context Contrastive Learning to align representations across lexical views, (ii) incorporates Ordinal Soft Contrastive Learning to preserve graded difficulty structure, and (iii) leverages complementary encoder models through Ridge ensembling. Together, these components enable more robust, aligned, and interpretable difficulty estimation.

\section{Background}

Recent studies show that transformer-based models can predict lexical difficulty from contextualized and multilingual representations \citep{shardlow-etal-2021-semeval, shardlow-etal-2024-bea, skidmore-etal-2025-transformer}. These approaches are motivated by the growing importance of lexical complexity prediction in applications such as text simplification, readability assessment, and language learning \citep{shardlow-2022-agree, rotaru-2021-andi}. In particular, shared tasks such as SemEval-2021 have demonstrated that fine-tuned transformer models can achieve strong performance by leveraging contextual information \citep{shardlow-etal-2021-semeval, rotaru-2021-andi}.

However, most approaches rely on regression-only training, where supervision is given only through a scalar difficulty score. This can improve prediction accuracy, but it does not directly structure the representation space. As a result, items with similar difficulty may not be close in the latent space, while items with different difficulty levels may not be clearly separated. This limitation becomes more pronounced in multilingual settings, where models must capture both cross-lingual alignment and language-specific variation \citep{skidmore-etal-2025-transformer}.

Contrastive learning addresses this limitation by shaping representations through relations between examples \citep{pmlr-v119-chen20j, NEURIPS2020_d89a66c7}. This fits lexical difficulty prediction, where difficulty depends on the target word, context, and learner-specific variation. Recent contrastive regression methods define similarity by proximity in continuous target values rather than discrete labels \citep{zha2023rank, keramati2024conr}, making them suitable for ordinal difficulty scores. We therefore formulate L1-aware lexical difficulty prediction as a multi-objective problem that combines direct regression with auxiliary contrastive supervision.

\section{Context-Aligned Contrastive Regression}
Building on recent work in lexical difficulty prediction and L1-aware modeling \citep{shardlow-etal-2021-semeval, shardlow-etal-2024-bea, skidmore-etal-2025-transformer}, we propose \textit{Context-Aligned Contrastive Regression}, which integrates direct regression with representation-level contrastive regularization. Given an input instance enriched with L1-aware contextual information, such as translated context and English target information, the model encodes multiple contextual views into a shared representation space. The resulting representation is used for both difficulty prediction and contrastive learning.
\begin{figure*}[t]
  \includegraphics[width=1.0\linewidth]{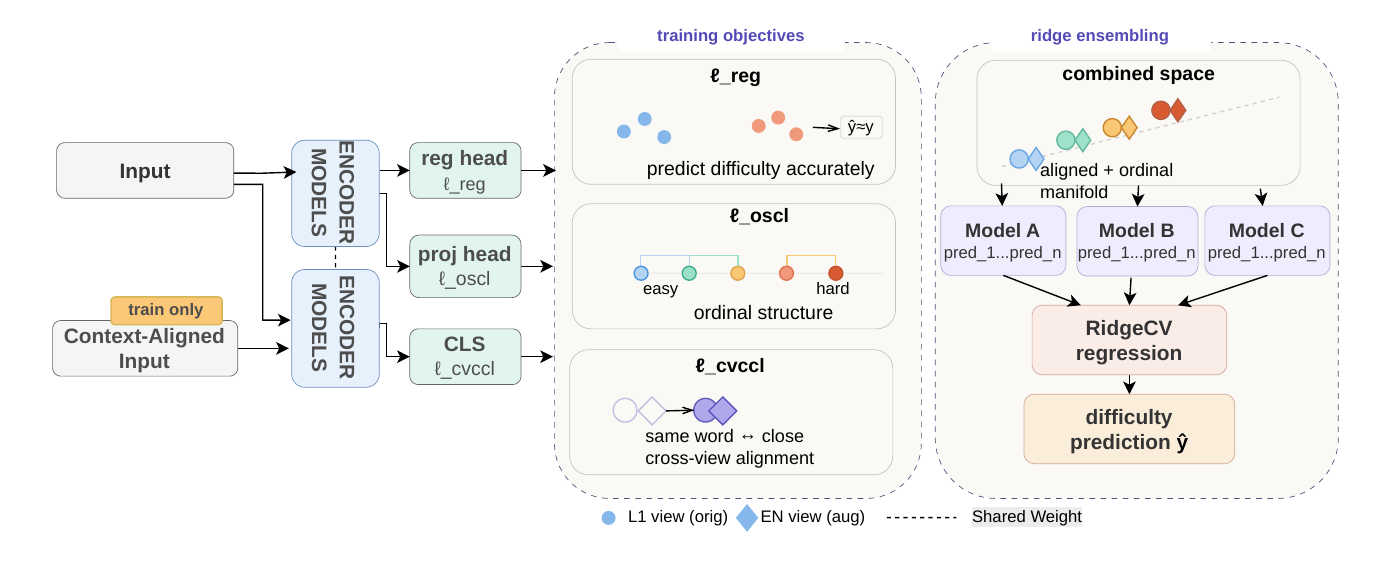}
  \caption {The proposed method combines regression and contrastive auxiliary objective, including cross-view alignment and ordinal-aware representation learning, to learn representations that are aligned across views and difficulty levels. Outputs from multiple encoder models are ensembled using ridge regression to produce the final prediction. 
  }
  \label{flowchart}
\end{figure*}

Specifically, we optimize the model using Regression loss (\S\ref{sec:regression_loss}) for direct difficulty prediction, a Cross-View Context Contrastive loss (\S\ref{sec:cross_view_loss}) for learning view-invariant contextual representations, and an Ordinal Soft Contrastive loss (\S\ref{sec:ordinal_soft_loss}) for encoding the continuous ordering of difficulty scores in the latent space. By combining this, the overall training objective does not merely fit the target score, but also provides a representation space that is both contextually stable and smoothly aligned with the ordinal structure of the task. The general explanation of our system is presented in the Figure \ref{flowchart}.

\subsection{Regression Objective}
\label{sec:regression_loss}
The regression objective is used to supervise the final prediction directly. Let $h_i \in \mathbb{R}^d$ denote the shared encoder representation of item $i$, and let $\hat{y}_i$ denote the scalar prediction generated by the regression head. The model is optimized with the mean squared error:
\begin{equation}
\mathcal{L}_{\mathrm{reg}}
=
\frac{1}{B}
\sum_{i=1}^{B}
(\hat{y}_i - y_i)^2,
\end{equation}
where $y_i$ is the label associated with item $i$ and $B$ is the batch size.

While this objective provides direct supervision for score prediction, it does not explicitly enforce representation consistency across alternative contextual realizations of the same lexical item, nor does it preserve proximity structure among items with similar difficulty levels. To address these limitations, we introduce two auxiliary contrastive objectives that regularize the representation space with respect to contextual alignment and ordinal difficulty structure.

\subsection{Cross-View Context Contrastive Objective}
\label{sec:cross_view_loss}

Lexical Difficulty is largely determined by intrinsic properties of the target word rather than by superficial variation in context \citep{paetzold-specia-2016-inferring, gooding-kochmar-2018-camb, shardlow-etal-2021-semeval}. However, contextualized encoders can mix lexical information with context-specific cues, which may produce inconsistent representations for the same lexical item across different contexts.

To reduce this effect, we use a cross-view contrastive objective for lexical difficulty prediction. Unlike standard contrastive learning methods that rely on stochastic augmentations \citep{oord2019representationlearningcontrastivepredictive, pmlr-v119-chen20j}, our method uses task-specific paired views derived from the L1-aware input and the context-aligned representation described in Section~\ref{sec:l1_aware_input}. This encourages the encoder to learn representations that remain stable across contextual variation while still distinguishing different lexical items \citep{NEURIPS2020_d89a66c7}.

\begin{equation}
\label{eq:tgt}
z_i^{\mathrm{tgt}} = H(en\_tgt_i)
\end{equation}

\begin{equation}
\label{eq:full}
z_i^{\mathrm{full}} = H(full\_input_i)
\end{equation}

Given a mini-batch of size $B$, each instance is encoded through two matched views, $en\_tgt$ (Eq.~\ref{eq:tgt}) and $full\_input$ (Eq.~\ref{eq:full}), yielding $2B$ representations. Since lexical difficulty prediction is formulated as a regression task, positive pairs are defined by cross-view correspondence rather than shared class labels. Thus, each anchor representation $z_i$ is paired with its matched representation from the alternative view. The objective is formalized as

{\small
\begin{equation}
\label{eq:cvccl}
\mathcal{L}_{\mathrm{CVCCL}}
=
-\frac{1}{2B}
\sum_{i=1}^{2B}
\log
\frac{
\exp\left(z_i^\top z_{i^{+}}/\tau\right)
}{
\sum_{k=1}^{2B}\mathbf{1}_{[k\neq i]}\exp\left(z_i^\top z_k/\tau\right)
}
\end{equation}}

\subsection{Ordinal Soft Contrastive Objective}
\label{sec:ordinal_soft_loss}

While the cross-view objective promotes consistency across views, it does not explicitly capture the ordinal structure of lexical difficulty. Since difficulty is represented as a continuous psychometric estimate rather than a discrete class label \citep{shardlow-etal-2021-semeval}, items with nearby scores should be closer in representation space than items with distant scores. Standard contrastive objectives rely on instance discrimination or discrete class supervision \citep{pmlr-v119-chen20j, NEURIPS2020_d89a66c7}, while recent regression-oriented contrastive methods address this limitation by organizing representations according to target distance or order \citep{zha2023rank, keramati2024conr, xue2024tractoscr}.

Following this motivation, we introduce an ordinal soft contrastive objective that replaces binary pair assignments with continuous pairwise weights derived from score proximity. This allows the model to preserve the graded structure of lexical difficulty in the embedding space. Let $u_i$ denote the representation of item $i$, and let $y_i$ be its lexical difficulty score. For a batch of size $B$, we define the affinity between items $i$ and $j$ as

\begin{equation}
w_{ij}
=
\exp\left(
-\frac{(y_i-y_j)^2}{2\sigma^2}
\right),
\qquad w_{ii}=0,
\end{equation}

where $\sigma$ controls how strongly the objective responds to differences in lexical difficulty. Pairs with similar scores receive larger weights, while pairs with distant scores receive smaller weights. We then define the similarity distribution for anchor $u_i$ over the remaining items in the batch as

\begin{equation}
p_{ij}
=
\frac{
\exp\left(\operatorname{sim}(u_i,u_j)/\tau\right)
}{
\sum_{k \neq i}
\exp\left(\operatorname{sim}(u_i,u_k)/\tau\right)
},
\qquad j \neq i,
\end{equation}

where $\operatorname{sim}(u_i,u_j)$ denotes cosine similarity and $\tau$ is a temperature parameter. The ordinal soft contrastive loss is then defined as

\begin{equation}
\mathcal{L}_{\mathrm{OSCL}}
=
\frac{1}{B}
\sum_{i=1}^{B}
\left(
-
\frac{
\sum_{j \neq i} w_{ij}\log p_{ij}
}{
\sum_{j \neq i} w_{ij}
}
\right).
\end{equation}

This objective complements the cross-view contrastive loss by shaping the representation space according to graded difficulty similarity. Items with nearby difficulty scores are encouraged to lie closer together, while items with larger score differences exert weaker attractive force. As a result, the learned embedding space reflects the continuous and ordinal nature of lexical difficulty more faithfully.

\subsection{Ridge-Based Ensemble}
\label{sec:ridge_ensemble}

After training the model with the proposed multi-objective learning framework, we further enhance prediction performance by leveraging an ensemble of models with different encoder backbones. Each model is trained independently using the same objective, resulting in diverse yet complementary predictions.

Let $f_k(x)$ denote the prediction of the $k$-th model for an input $x$. We construct a meta-representation by stacking predictions from $K$ models:
\begin{equation}
\mathbf{z}(x) = [f_1(x), f_2(x), \dots, f_K(x)].
\end{equation}
We then apply ridge regression to combine these predictions, allowing the model to learn adaptive weights over individual predictors while maintaining robustness through regularization. This approach provides a more flexible alternative to simple averaging and leads to more stable and accurate predictions.

\section{Experimental Setup}
\subsection{Dataset}
We perform lexical difficulty prediction as a regression task using the multilingual L1-aware dataset introduced by \citet{skidmore-etal-2025-transformer}. The data are organized into three first-language groups, namely German, Spanish, and Mandarin Chinese. Total of each L1 group is presented in Table \ref{tab:data_distribution}. For the translated training and development splits, we additionally include \textit{en\_context}, as described in Section~\ref{sec:l1_aware_input}. 

Furthermore, the target variable, \textit{GLMM\_score}, represents the estimated difficulty of each vocabulary item. It is computed from large-scale learner response data using a generalized linear mixed model (GLMM) framework, which models both item-level and learner-level variability \citep{schmitt2024knowledge}. Lower scores indicate greater lexical difficulty.

\begin{table}[h]
\centering
\caption{Data distribution across L1 groups and splits.}
\label{tab:data_distribution}
\begin{tabular}{lrrr}
\hline
\textbf{L1 group} & \textbf{Train} & \textbf{Dev} & \textbf{Test} \\
\hline
German            & 6,091 & 677 & 748 \\
Spanish           & 6,091 & 677 & 748 \\
Mandarin Chinese  & 6,091 & 677 & 748 \\
\hline
Total             & 18,273 & 2,031 & 2,244 \\
\hline
\end{tabular}
\end{table}

\subsection{L1-Aware Input Representation}
\label{sec:l1_aware_input}
\citet{skidmore-etal-2025-transformer} reported that their best-performing setup used a multilingual model trained jointly on all L1 subsets of the KVL together with an L1-aware input representation. Following this formulation, we represent each instance as a single sequence that combines the L1 source word ($w$), its L1 context ($ctx$), the English clue ($clue$), and the target English word ($tgt$). These components are separated by a special token before being passed to a pretrained encoder,  as shown in the example input text below:

\begin{tcolorbox}[colback=gray!15, colframe=gray!50, boxrule=0pt, arc=3pt]
\texttt{casa [SEP] Vivo en una casa grande que
tiene tres dormitorios. [SEP] h\_\_\_\_ [SEP] house
}
\end{tcolorbox}
In addition, we construct a context-aligned view by combining the English target word from the source language input ($en\_tgt$) with the translated English context ($en\_ctx$), which is obtained by translating the source-language context $ctx$.\footnote{We use \texttt{google/mt5-large} to translate the source-language context.} This augmented view is then used to form contrastive pairs for cvccl objective (\S\ref{sec:cross_view_loss}). 


\subsection{Model Architecture}
Our model follows a shared-encoder architecture with task-specific heads for regression and contrastive learning. Given an input sequence, the encoder produces a contextualized representation from the \texttt{[CLS]} token, which is then used for both prediction and representation-level objectives.


\paragraph{Projection Head}
For contrastive learning, we apply a projection head that maps the encoder representation into a lower-dimensional space. The projected representation is used for contrastive objectives. Separating the projection space from the original representation allows the model to preserve task-relevant features for regression while learning more structured representations for contrastive objectives.

\paragraph{Models Setting}
We use three multilingual encoder models as base learners in the ensemble: XLM-RoBERTa \citep{conneau-etal-2020-unsupervised}, multilingual DeBERTaV3 \citep{he2023debertav3improvingdebertausing}, and mmBERT \citep{marone2025mmbertmodernmultilingualencoder}. At inference time, predictions from all base learners are combined using a ridge regression meta-model, which learns to aggregate their outputs into a final prediction. The hyperparameters and training settings are reported in Appendix~\ref{sec:training_setting}.




\section{Results \& Discussion}
In this section, we present our main result (see Table \ref{tab:ensemble_model_results} and Appendix \ref{sec:leaderboard_comparison} for detailed results) along with its observations and discussion. We also provide qualitative examples of analysis across languages, describing the top-5 improved and failure cases in the dataset.

\subsection{Main Result}
\label{sec:main_finding}

\begin{table*}[!htbp]
\centering
\resizebox{\linewidth}{!}{
\begin{tabular}{l ccccccccc}
\toprule
\multirow{2}{*}{\textbf{Model}} &
\multicolumn{3}{c}{\textbf{ES}} &
\multicolumn{3}{c}{\textbf{DE}} &
\multicolumn{3}{c}{\textbf{CN}} \\
\cmidrule(lr){2-4} \cmidrule(lr){5-7} \cmidrule(lr){8-10}
& RMSE $\downarrow$ & MSE $\downarrow$ & $\rho$ $\uparrow$
& RMSE $\downarrow$ & MSE $\downarrow$ & $\rho$ $\uparrow$
& RMSE $\downarrow$ & MSE $\downarrow$ & $\rho$ $\uparrow$ \\
\toprule

\multicolumn{10}{c}{\textbf{Individual Model}} \\
\midrule
XLM-R (base) 
& 1.257 & 1.580 & 0.765 
& 1.258 & 1.583 & 0.773 
& 1.140 & 1.300 & 0.753 \\

mDeBERTa-v3 (base)
& 1.151 & 1.326 & 0.836
& 1.160 & 1.344 & 0.833
& 1.027 & 1.054 & 0.855 \\

mmBERT (base)
& 1.079 & 1.164 & 0.825
& 1.010 & 1.021 & 0.826
& 0.911 & 0.830 & 0.843 \\
\midrule
\multicolumn{10}{c}{\textbf{Ensemble Variants}} \\
\midrule
ensemble baseline 
& 1.051 & 1.104 & 0.835
& 0.998 & 0.996 & 0.831
& 0.893 & 0.798 & 0.848 \\

full\_input (CVCCL + OSCL)
& 1.045 & 1.092 & 0.835 
& 1.016 & 1.032 & 0.828 
& 0.907 & 0.823 & 0.843 \\

en\_ctx + en\_tgt (CVCCL + OSCL)
& 1.052 & 1.107 & 0.831 
& 1.004 & 1.008 & 0.829 
& 0.892 & 0.796 & 0.846 \\

en\_ctx (CVCCL + OSCL)
& \textbf{1.041} & \textbf{1.084} & \textbf{0.837} 
& 1.010 & 1.020 & 0.827 
& 0.890 & 0.792 & 0.852 \\

en\_tgt (CVCCL + OSCL)
& 1.063 & 1.130 & 0.826 
& \textbf{0.997} & \textbf{0.994} & \textbf{0.832} 
& \textbf{0.880} & \textbf{0.774} & \textbf{0.853} \\

en\_tgt (CVCCL) 
& 1.052 & 1.107 & 0.831 
& 0.998 & 0.996 & 0.831 
& 0.891 & 0.794 & 0.847 \\

en\_tgt (OSCL) 
& 1.077 & 1.160 & 0.825 
& 1.001 & 1.002 & 0.832 
& 0.866 & 0.750 & 0.858 \\

\bottomrule
\end{tabular}
}
\caption{Performance comparison with RMSE, MSE, and Spearman's $\rho$ across languages and models in the BEA shared task test set. Individual model denotes fine-tuned model with regression head on top of the architecture and \textit{full\_input} or L1-aware input representation \S\ref{sec:l1_aware_input} as model input. Lower RMSE/MSE and higher $\rho$ indicate better performance.}
\label{tab:ensemble_model_results}
\end{table*}


\paragraph{Performance Across Languages}

Table~\ref{tab:ensemble_model_results} shows that the ensemble consistently outperforms individual base models across the three L1 groups. These gains suggest that combining input design, auxiliary objectives, and model ensembling provides more robust predictions than relying on a single encoder. Spanish achieves the best performance with the $en\_ctx$ setting, while German and Chinese benefit more from $en\_tgt$-based representations. Rather than indicating instability, this pattern suggests that lexical difficulty is shaped by different linguistic cues across L1 backgrounds. This makes the variation between input settings analytically meaningful and motivates a deeper examination of model behavior and representation structure. \footnote{The Ensemble Baseline combines the XLM-R, mDeBERTa-v3, and mmBERT baseline models using ridge regression.}

\paragraph{Effect of Input and Objective Design}    
A closer comparison of input and objective variants shows that more complex configurations do not always improve performance. Although $full\_input$ (\S\ref{sec:l1_aware_input}) and $en\_ctx + en\_tgt$ perform strongly, they do not consistently outperform focused single-view representations, such as \textit{en\_tgt (CVCCL + OSCL)} or \textit{en\_context (CVCCL + OSCL)}. This suggests that carefully selected inputs can be more effective than combining all available features, likely by reducing noise and improving representation stability.

At the objective level, CVCCL + OSCL remains consistently competitive across languages. It does not always achieve the best score, but it stays among the top configurations in each setting. This indicates that contrastive alignment and ordinal supervision provide complementary learning signals, while their effectiveness depends on the structure of the input information.



\paragraph{BEA 2026 Submission}
We submitted our best model to the BEA 2026 Shared Task Closed Track. Full leaderboard comparisons are provided in Appendix~\ref{sec:leaderboard_comparison}. Our submission, \textbf{TOEBM}, achieved top-15 performance across all L1 groups, ranking 14th for Spanish, 11th for German, and 7th for Mandarin Chinese, as shown in Tables~\ref{tab:leaderboard-st1}, \ref{tab:leaderboard-st2}, and \ref{tab:leaderboard-st3}, respectively. This pattern suggests that the proposed approach remains reasonably robust across different L1 backgrounds, even when the relative difficulty cues vary by language. Although its RMSE is about $\pm0.125$ higher than the top-ranked Glite Team on average, the model maintains competitive Pearson correlation scores across languages, with 0.832 for Spanish and German, and 0.853 for Mandarin Chinese. These results indicate that, while the model is not yet fully optimized for minimizing prediction error, it captures the relative ordering of lexical difficulty effectively.



\subsection{Representation Analysis}
\begin{figure}[ht]
    \centering
    
    \begin{subfigure}[t]{0.48\linewidth}
        \centering
        \includegraphics[width=\linewidth]{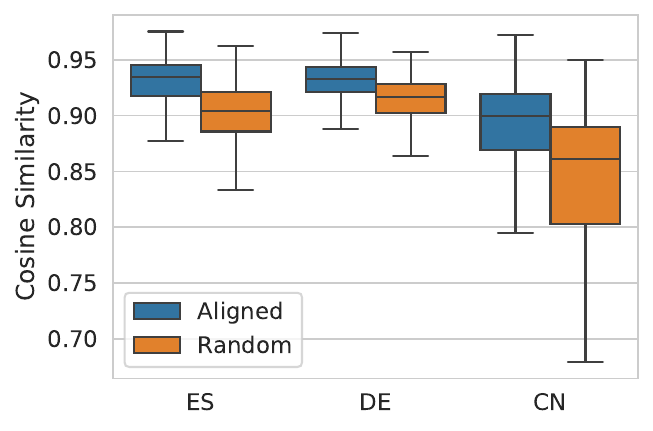}
        \caption{Cross-lingual alignment}
        \label{fig:alignment}
    \end{subfigure}
    \hfill
    \begin{subfigure}[t]{0.48\linewidth}
        \centering
        \includegraphics[width=\linewidth]{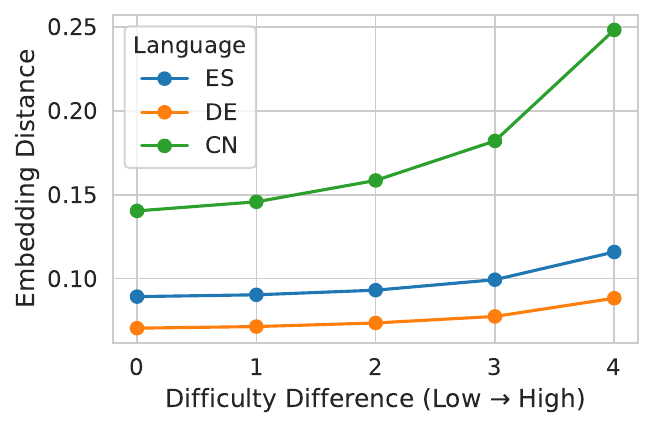}
        \caption{Ordinal structure}
        \label{fig:ordinal}
    \end{subfigure}

    \caption{
    Representation analysis across Spanish, German, and Chinese using mmBERT-base. 
    (a) Cosine similarity between aligned cross-lingual lexical pairs and randomly paired inputs.
    (b) Relationship between embedding distance and absolute lexical difficulty score differences, larger score gaps correspond to greater representational separation.
    }
    \label{fig:representation_analysis}
    
    \vspace{-0.8em}
\end{figure}

To better understand the impact of the contrastive objectives, we analyze the learned representations beyond the final prediction layer. In particular, we examine whether the contrastive losses improve cross-lingual alignment (\S\ref{sec:cross_view_loss}) and preserve the ordinal structure of lexical difficulty (\S\ref{sec:ordinal_soft_loss}) across multiple languages.

\paragraph{Cross-lingual Representation Alignment}
We first examine whether the contrastive objective leads to improved cross-lingual alignment in the learned representations. To this end, we measure cosine similarity between representations of semantically equivalent inputs (L1 and English) and compare them against randomly paired inputs.\footnote{The details of measurement technique is presented in Appendix \ref{sec:cross_lingual_details}.} As shown in Figure~\ref{fig:alignment}, aligned pairs consistently exhibit higher similarity than random pairs across all three languages (ES, DE, and CN), indicating that the model learns a shared semantic space across languages.

Notably, the degree of separation varies across languages, with German showing the clearest gap, while Chinese exhibits greater variance and overlap. The presence of overlap between the two distributions suggests that the alignment is not fully separable. This indicates that while the contrastive objective contributes to reducing cross-lingual representation gaps, it does not enforce strict alignment. We hypothesize that this moderate alignment is beneficial, as it preserves language-specific nuances that may still be relevant for lexical difficulty prediction.

\paragraph{Ordinal Structure in Representation}
We next analyze whether the learned representations capture the ordinal nature of lexical difficulty. Specifically, we examine the relationship between embedding distances and differences in GLMM scores\footnote{The details of measurement technique is presented in Appendix \ref{sec:ordinal_structure}}. As shown in Figure~\ref{fig:ordinal}, we observe a clear and consistent increasing trend across all three languages, where pairs of instances with larger difficulty differences exhibit greater distances in the embedding space.

We further quantify this trend using Spearman’s rank correlation. We observe a consistent positive monotonic relationship across all three languages (Spearman’s $\rho$ = 0.22 for ES, 0.20 for DE, and 0.39 for CN), indicating that embedding distances tend to increase with larger differences in lexical difficulty. While the correlation is moderate in magnitude, its consistency across languages suggests that the ordinal structure is reliably encoded in the learned representations, although the relationship is not strictly linear.


\paragraph{Input Representation Behavior}

\begin{table}[t]
\centering
\small
\begin{tabular}{lccc}
\toprule
\textbf{Condition} & \textbf{Low} & \textbf{High} & \textbf{Rel. $\downarrow$ (\%)} \\
\midrule
Context Length      & 0.841 & 0.834 & 0.8 \\
Lexical Diversity   & 0.847 & 0.837 & 1.2 \\
Target Word Length         & 0.843 & 0.831 & 1.4 \\
Orthographic Complexity         & 0.855 & 0.813 & 4.9 \\
\bottomrule
\end{tabular}
\caption{Mean absolute error (MAE) of the Ridge ensemble across input characteristics on the Spanish dataset. Relative improvement (\%) measures the reduction from low to high condition. Lower values indicate better performance.}
\label{tab:input_sensitivity}
\vspace{-0.8em}
\end{table}

Beyond representation structure, we further investigate how input characteristics influence prediction behavior. Specifically, we analyze prediction error with respect to several surface-level input properties, including context length, lexical diversity (measured via type-token ratio), target word length, and orthographic complexity. We observe a generally consistent pattern in which richer contextual and lexical signals are associated with lower prediction error. In particular, longer and lexically more diverse contexts, as well as more orthographically informative target words, tend to improve prediction quality, suggesting that contextual and lexical richness provide useful signals for lexical difficulty modeling.

However, the strength of this effect varies across languages. As shown in Table~\ref{tab:input_sensitivity}, for Spanish, the improvements are more pronounced, especially in lexically diverse contexts and orthographically complex words, which exhibit the largest relative gain. This indicates that Spanish benefits more from contextualized representations, explaining why the $en\_ctx$ configuration outperforms target-only inputs in this setting. In contrast, for other languages, the gains from context are less dominant, suggesting that target-side information can be sufficient depending on the dataset characteristics. This finding is consistent with the representation-level analysis, where contextual signals contribute to richer and more structured embeddings, which in turn improve downstream prediction.

\subsection{Ensemble Investigation}
We analyze the contribution of each base model to the meta-input of the Ridge regression ensemble. This analysis aims to assess how well individual models approximate the GLMM scores and whether they provide complementary prediction patterns that can be effectively leveraged by the ensemble.

\begin{figure*}[t]
  \centering
  \includegraphics[width=0.8\linewidth]{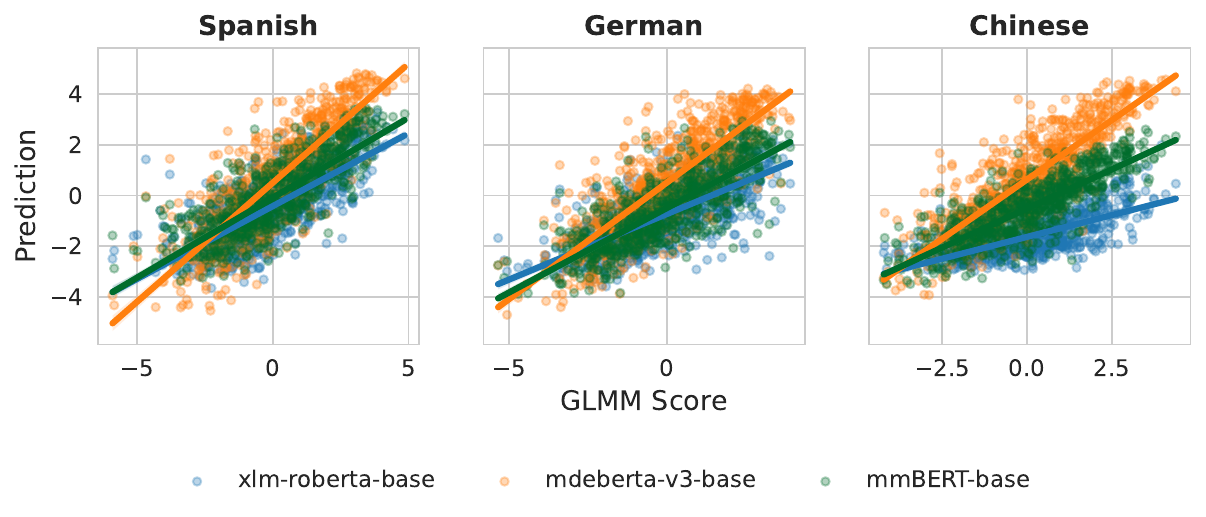}
  \caption{Relationship between predicted scores and ground-truth GLMM scores across Spanish (ES), German (DE), and Chinese (CN). Each subplot shows regression trends for individual base models, highlighting systematic differences in prediction behavior, including underestimation, overestimation, and balanced estimation patterns.}
  \label{fig:prediction_ground_truth}
\end{figure*}

\paragraph{Base Models Prediction Behaviour}
Figure~\ref{fig:prediction_ground_truth} illustrates the relationship between predicted scores and ground truth GLMM scores across all three L1 groups. The models exhibit systematic differences in how lexical difficulty is estimated. XLM-RoBERTa consistently produces a relatively flat trend, particularly in the Chinese dataset, indicating a tendency to compress predictions toward the mean and underestimate more difficult items. In contrast, mDeBERTa-v3 shows a steeper slope, often overshooting the ground-truth line, suggesting systematic overestimation for complex lexical items. Meanwhile, mmBERT demonstrates a more moderate slope that more closely follows the optimal diagonal trend, indicating a more balanced estimation behaviour.

\paragraph{Statistical Patterns}
These contrasting trends reveal that the base models capture different aspects of the lexical difficulty space. While XLM-RoBERTa tends to under-predict and mDeBERTa-v3 tends to over-predict, mmBERT produces more balanced predictions with errors closer to zero and generally competitive variance. This complementary behavior is further reflected in the trade-off between correlation and error metrics, where mDeBERTa-v3 often achieves higher Pearson correlation while mmBERT attains lower RMSE and MAE in several cases. 

When considered alongside the statistical results in Table~\ref{tab:base_model_analysis} (Appendix), these findings suggest that the models are not redundant but instead provide diverse and complementary signals. Such diversity is crucial for the effectiveness of the ensemble, as it enables the meta-learner to reconcile systematic biases and produce more accurate final predictions.

\begin{figure}[!h]
    \centering
    \includegraphics[width=0.65\linewidth]{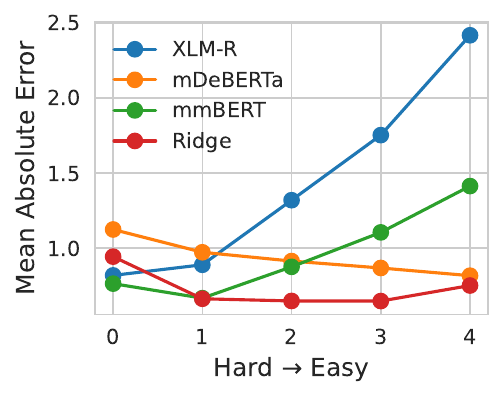}
    \caption{
Mean absolute error (MAE) across lexical difficulty bins, averaged over all three L1 groups. Difficulty bins are derived from GLMM score quantiles and ordered from harder (lower GLMM scores) to easier (higher GLMM scores) lexical items.
}
    \label{fig:difficulty_bin_error}
    \vspace{-0.8em}
\end{figure}

\paragraph{Difficulty-wise Error Analysis}
To further evaluate ensemble effectiveness across the lexical difficulty spectrum, we analyze model errors over five GLMM-based difficulty bins averaged across all three L1 groups. Figure~\ref{fig:difficulty_bin_error} presents the resulting mean absolute error (MAE) patterns.

We observe that XLM-RoBERTa exhibits a substantial increase in error from harder to easier lexical items, indicating that its predictions become progressively less reliable for comparatively easier words and suggesting strong score compression effects. Similarly, mmBERT demonstrates moderate but consistent error growth toward easier bins, although its overall performance remains more balanced than XLM-RoBERTa. In contrast, mDeBERTa-v3 maintains relatively stable performance across the full spectrum, with only minor variation between difficulty levels.

Importantly, the Ridge ensemble consistently achieves the lowest or near-lowest MAE across nearly all bins while maintaining the flattest error profile overall. This suggests that the meta-learner effectively reconciles the complementary weaknesses of individual base models, substantially reducing systematic prediction bias and producing more stable lexical difficulty estimation across both harder and easier lexical items.

\subsection{Improved vs Failures Cases}
To complement the quantitative findings in Section~\S\ref{sec:main_finding}, we further examine representative cases to understand how the ensemble model corrects prediction errors from individual base models. 

\begin{table}[t]
\centering
\small
\begin{tabular}{lcccc}
\toprule
\textbf{Lang} & \textbf{Word} & \textbf{GLMM} & \textbf{mmBERT} & \textbf{Ridge} \\
\midrule
ES & cruise     & 1.21 & -0.50 & 1.23 \\
ES & run        & 2.20 & 0.31  & 1.99 \\
ES & character  & 1.08 & -1.11 & 0.55 \\
\cellcolor{selected}ES & \cellcolor{selected}dining & \cellcolor{selected}-4.67 & \cellcolor{selected}-0.76 & \cellcolor{selected}0.19 \\
\cellcolor{selected}ES & \cellcolor{selected}baking & \cellcolor{selected}-3.78 & \cellcolor{selected}-0.20  & \cellcolor{selected}0.87 \\
\midrule
CN & schoolmate & 1.32 & -2.55 & -0.09 \\
CN & dating     & 0.64 & -2.31 & -0.001 \\
CN & goldfish   & 1.51 & -1.44 & 0.63 \\
\cellcolor{selected}CN & \cellcolor{selected}double \cellcolor{selected}& \cellcolor{selected}-2.58 & \cellcolor{selected}0.11  & \cellcolor{selected}0.82 \\
\cellcolor{selected}CN & \cellcolor{selected}stake & \cellcolor{selected}-4.19 & \cellcolor{selected}-1.67 & \cellcolor{selected}-1.13 \\
\midrule
DE & cruise     & 1.21 & -1.48 & 0.60 \\
DE & handbook   & 0.95 & -1.36 & 0.31 \\
DE & workbook   & 2.10 & -2.07 & -0.44 \\
\cellcolor{selected}DE & \cellcolor{selected}anger & \cellcolor{selected}-3.39 & \cellcolor{selected}0.10  & \cellcolor{selected}0.57 \\
\cellcolor{selected}DE & \cellcolor{selected}umbrella   & \cellcolor{selected}3.24 & \cellcolor{selected}-1.13  & \cellcolor{selected}-0.20 \\
\bottomrule
\end{tabular}
\caption{Qualitative examples across Spanish (ES), Chinese (CN), and German (DE), comparing mmBERT-base and Ridge ensemble predictions. The first three rows per language show cases where the ensemble corrects large underestimation errors from base models, while the last two rows highlight failure cases.}
\label{tab:qualitative_cases}
\vspace{-0.8em}
\end{table}

\paragraph{Error Correction}
Table~\ref{tab:qualitative_cases} presents selected examples across Spanish, Chinese, and German. We observe that the ensemble consistently corrects large underestimation errors from individual models. For instance, words such as \textit{cruise}, \textit{run}, and \textit{character} in Spanish, as well as \textit{schoolmate} and \textit{dating} in Chinese, are substantially underestimated by mmBERT, yet the Ridge ensemble successfully adjusts their predictions toward the ground-truth GLMM scores. Similar correction patterns are observed in German, suggesting that the ensemble effectively leverages complementary signals across models to reduce systematic bias.

\paragraph{Failure Cases}
Despite these substantial improvements, certain lexical items remain challenging even after ensembling. In several cases, such as \textit{dining} and \textit{baking} in Spanish, \textit{double} and \textit{stake} in Chinese, and \textit{anger} and \textit{umbrella} in German, the predicted difficulty remains far from the ground-truth values. These errors often involve strong overestimation or residual bias, indicating that certain lexical items remain challenging due to semantic ambiguity or context-dependent interpretations. This suggests that while the ensemble mitigates systematic biases, it cannot fully resolve cases where all base models fail to capture the underlying difficulty signal.

\section{Conclusion}
Our method consistently outperforms individual multilingual encoders across Spanish, German, and Chinese learner groups, effectively correcting systematic under- and overestimation biases while maintaining stable performance across all difficulty levels. Representation analysis confirms that the contrastive objectives enhance cross-lingual alignment and preserve the ordinal structure of lexical difficulty in the embedding space. Collectively, these results demonstrate that combining direct regression with representation-level regularization significantly improves both the accuracy and structural reliability of L1-aware lexical difficulty prediction.

\section*{Acknowledgments}
The authors gratefully acknowledge Mantera Studio for providing the computational resources and APIs utilized in this research. Additionally, this work was partially supported by the 2026 Publication Funding from the Department of Computer Science and Electronics at Universitas Gadjah Mada. 

\section*{Limitations}
Although the proposed method improves performance across the evaluated L1 groups, several limitations remain. First, the experiments are limited to Spanish, German, and Mandarin Chinese, so the findings may not fully generalize to other L1 backgrounds. Second, the use of contrastive objectives introduces additional training objectives and weighting hyperparameters, which may increase sensitivity to objective balancing across languages. Third, the ridge ensemble improves predictive performance by combining multiple encoders, but it also increases computational cost because several models must be trained and evaluated. Future work should examine more robust weighting strategies, reduce the computational overhead of ensembling, and evaluate the method on a wider range of L1 backgrounds.

\section*{Ethical Considerations}
We acknowledge that our research utilized AI tools for writing, rewriting, and generating code. Although these tools offer significant advantages in terms of efficiency and productivity, their use raises important ethical considerations. We recognize the potential for bias and errors inherent in AI-generated content and have taken steps to mitigate these risks through rigorous human review and validation. Furthermore, we are mindful of the potential impact on the broader software development community, particularly regarding job displacement and the need for upskilling. We believe that responsible AI integration should prioritize transparency, accountability, and the empowerment of human developers, ensuring that these tools augment rather than replace human expertise. This research aims to contribute to the ongoing dialogue on ethical AI development and usage, advocating for a future where AI tools are harnessed responsibly to enhance human creativity and innovation in the field of software engineering.
\bibliography{custom}

\appendix

\section{BEA 2026 Leaderboard Comparison}
\label{sec:leaderboard_comparison}

Table~\ref{tab:leaderboard-st1}, Table~\ref{tab:leaderboard-st2}, and Table~\ref{tab:leaderboard-st3} present selected leaderboard comparisons for the Spanish, German, and Chinese subtasks, respectively. Rather than listing the complete leaderboard, we report the highest-ranked systems and nearby reference systems to contextualize our submission. Our team is registered as \textbf{TOEBM}. In the closed-track results, TOEBM ranked 14th for Spanish, 11th for German, and 7th for Chinese, with RMSE scores of 1.063, 0.997, and 0.880, respectively. The leaderboard scores differ slightly from those reported in the main experimental section because the present paper uses a unified prompt setting across all subtasks, while the official leaderboard reflects the submitted prediction files. Nevertheless, the relative ranking of our submission remains unchanged.

\begin{table}[!ht]
    \centering
    \resizebox{\linewidth}{!}{
    \begin{tabular}{lcccc}
        \toprule
        \textbf{\#} & \textbf{Team Name} & \textbf{Prediction Type} & \textbf{RMSE} & \textbf{Pearson} \\
        \midrule
        1 & Glite & predictions\_run\_3 & 0.903 & 0.877 \\
        ... & ... & ... & ... & ... \\
        7 & Sakura  & predictions\_closed\_max & 0.983 & 0.854  \\
        11 & NLP-Explorers & predictions\_run\_3 & 1.041 & 0.838  \\
        ... & ... & ... & ... & ... \\
        \cellcolor{oursgray}14 & \cellcolor{oursgray} TOEBM & \cellcolor{oursgray} predictions\_run\_3 & \cellcolor{oursgray} 1.063 & \cellcolor{oursgray} 0.826 \\
        \bottomrule
    \end{tabular}
    }
    \caption{Leaderboard comparison of Closed Track Spanish Language}
    \label{tab:leaderboard-st1}
\end{table}

\begin{table}[!ht]
    \centering
    \resizebox{\linewidth}{!}{
    \begin{tabular}{lcccc}
        \toprule
        \textbf{\#} & \textbf{Team Name} & \textbf{Prediction Type} & \textbf{RMSE} & \textbf{Pearson} \\
        \midrule
        1 & Glite & predictions\_run\_3 & 0.885 & 0.871 \\
        ... & ... & ... & ... & ... \\
        4 & Uogal  & 8enc\_dmeta\_elasticnet...csv & 0.903 & 0.869  \\
        7 & NLP-Explorers & predictions\_run\_3 & 0.992 & 0.845  \\
        ... & ... & ... & ... & ... \\
        \cellcolor{oursgray}
        11 & \cellcolor{oursgray} TOEBM & \cellcolor{oursgray} predictions\_run\_3 & \cellcolor{oursgray} 0.997 & \cellcolor{oursgray} 0.832 \\
        \bottomrule
    \end{tabular}
    }
    \caption{Leaderboard comparison of Closed Track German Language}
    \label{tab:leaderboard-st2}
\end{table}

\begin{table}[!ht]
    \centering
    \resizebox{\linewidth}{!}{
    \begin{tabular}{lcccc}
        \toprule
        \textbf{\#} & \textbf{Team Name} & \textbf{Prediction Type} & \textbf{RMSE} & \textbf{Pearson} \\
        \midrule
        1 & Glite & predictions\_run\_3 & 0.776 &0.889 \\
        ... & ... & ... & ... & ... \\
        4 & Sakura  & predictions\_closed\_max &0.816 & 0.874  \\
        5 & uogal & 8enc\_dmeta\_elasticnet...csv & 0.820 & 0.879  \\
        ... & ... & ... & ... & ... \\
        \cellcolor{oursgray}7& \cellcolor{oursgray} TOEBM & \cellcolor{oursgray} predictions\_run\_3 & \cellcolor{oursgray} 0.880 & \cellcolor{oursgray} 0.853 \\
        \bottomrule
    \end{tabular}
    }
    \caption{Leaderboard comparison of Closed Track Chinese Language}
    \label{tab:leaderboard-st3}
\end{table}

\section{Base Model Statistic}
\begin{table*}[t]
\centering
\small
\begin{tabular}{lccccccccc}
\hline
\textbf{Metrics} & \multicolumn{3}{c}{\textbf{Español}} & \multicolumn{3}{c}{\textbf{Deutsch}} & \multicolumn{3}{c}{\textbf{Chinese}} \\
\cline{2-10}
 & \textbf{XLM-R} & \textbf{mDeBERTa} & \textbf{mmBERT} 
 & \textbf{XLM-R} & \textbf{mDeBERTa} & \textbf{mmBERT} 
 & \textbf{XLM-R} & \textbf{mDeBERTa} & \textbf{mmBERT} \\
\hline
RMSE & 1.326 & 1.266 & \textbf{1.113} & 1.441 & \textbf{1.219} & 1.154 & 2.070 & \textbf{1.135} & 1.091 \\
MAE & 1.066 & 0.993 & \textbf{0.861} & 1.145 & \textbf{0.959} & 0.908 & 1.793 & \textbf{0.891} & 0.880 \\
Pearson & 0.746 & \textbf{0.836} & 0.808 & 0.719 & 0.832 & \textbf{0.813} & 0.689 & 0.858 & \textbf{0.824} \\
Mean Error & -0.432 & \textbf{0.497} & -0.098 & -0.722 & \textbf{0.507} & -0.491  & -1.644 & \textbf{0.628} & -0.520 \\
Std Error & 1.253 & 1.165 & \textbf{1.109} & 1.247 & 1.108 & 1.045 & 1.258 & 0.945 & \textbf{0.960} \\
\hline
\end{tabular}
\caption{Performance and error analysis of base models across three L1 groups. Best values per language are highlighted in bold.}
\label{tab:base_model_analysis}
\end{table*}

\section{Representation analysis}
To analyze the learned representation space, we distinguish between the Ridge ensemble used for prediction and the fused embedding used for visualization. The Ridge ensemble prediction is defined as
\begin{equation}
\hat{y}(x) = b + \sum_{n=1}^{N} w_n f_n(x),
\end{equation}
where $f_n(x)$ denotes the scalar prediction produced by the $n$-th base model, $w_n$ is the learned Ridge coefficient for that model, and $b$ is the intercept. This formulation is used only for final score prediction. It is not suitable for direct t-SNE visualization because it produces a one-dimensional output.

For representation analysis, we instead construct a fused embedding by concatenating the weighted embedding vectors from each base model:
\begin{equation}
z(x)=\left[w_1E_1(x);\,w_2E_2(x);\,\cdots;\,w_NE_N(x)\right],
\end{equation}
where $E_n(x)\in\mathbb{R}^{d_n}$ is the embedding produced by the $n$-th encoder. In this study, $E_n(x)$ is obtained from the CVCCL+OSCL model using the $en\_tgt$ view as the encoder input. The Ridge coefficients are then used as model-level weights to scale each embedding before concatenation.

We apply t-SNE to the fused representation $z(x)$ to visualize the structure of the embedding space. The resulting points are grouped into five difficulty bins based on the minimum and maximum GLMM scores. This allows us to inspect whether examples with similar lexical difficulty are located close to one another in the representation space, while the Ridge ensemble prediction $\hat{y}(x)$ is used separately to evaluate predictive performance.
\begin{figure*}[ht]
    \centering
    \includegraphics[width=\linewidth]{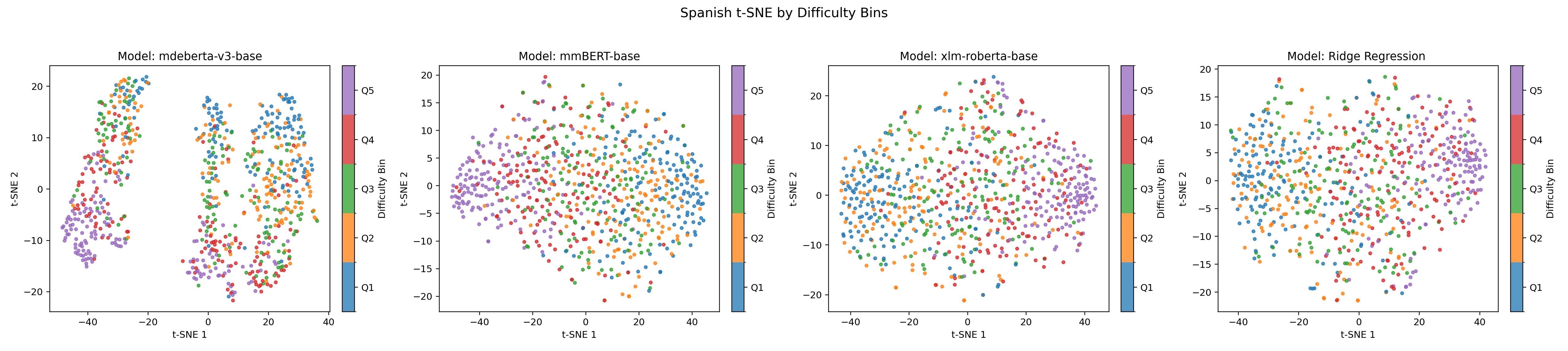}
    \includegraphics[width=\linewidth]{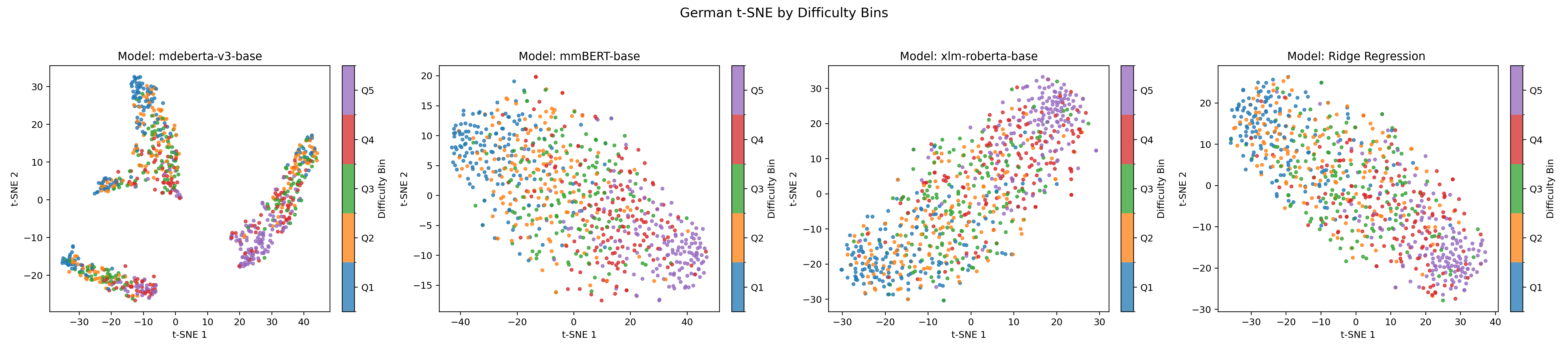}
    \includegraphics[width=\linewidth]{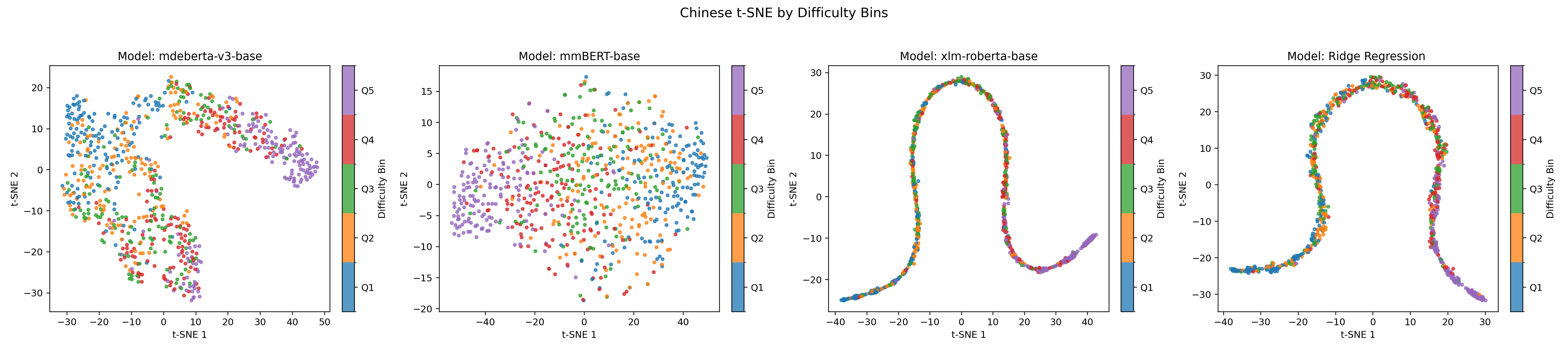}
    \caption{t-SNE visualization of fused model representations for Spanish, German, and Chinese. Each point represents one sample, and colors indicate five GLMM difficulty bins constructed from the minimum and maximum GLMM scores. The visualization is based on the weighted concatenation of encoder embeddings, while $\hat{y}(x)$ is used separately for predictive evaluation.}
    \label{fig:tsne_truebin_panel}
\end{figure*}
Figure~\ref{fig:tsne_truebin_panel} shows that the separation across difficulty bins is generally strong, although the quality of the structure varies across models and languages. Since GLMM is a continuous score and the bins are created only for visualization, perfect discrete clusters are not expected. Even so, a clear ordinal organization is visible in many panels, where neighboring regions are dominated by adjacent difficulty bins and the progression from lower to higher difficulty appears smooth.

Among the individual encoders, \texttt{mdeberta-v3-base} often produces well-separated local clusters, especially in Spanish and German, indicating that the learned space captures meaningful difficulty structure. \texttt{mmBERT-base} also shows some ordering, but the bins are more mixed and the boundaries between difficulty levels are less distinct. In contrast, \texttt{xlm-roberta-base} provides the clearest structure for Mandarin Chinese. In the Chinese panel, the bins are arranged along a highly organized manifold with a very strong gradual transition from easier to harder examples. This suggests that \texttt{xlm-roberta-base} captures the ordinal nature of lexical difficulty particularly well for Mandarin Chinese.

The Ridge-weighted fused representation remains competitive and in some cases preserves a smooth global arrangement of the bins, showing that the ensemble combines complementary information from the base encoders. However, the clearest single-model structure for Mandarin Chinese is produced by \texttt{xlm-roberta-base}. Overall, these visualizations support the claim that the CVCCL+OSCL representations encode lexical difficulty in a meaningful way, with strong separation between bins and especially clear ordinal structure in the Mandarin Chinese setting.
\section{Training Settings}
\label{sec:training_setting}
All models are trained using the Adam optimizer for 5 epochs with a learning rate of \texttt{2e-5}, a batch size of \texttt{8}, weight decay of \texttt{0.01}, and a warmup ratio of \texttt{0.06}. The best model is selected based on the Mean Squared Error (MSE) on the validation set. The maximum input length is set to 128 tokens, while the maximum length for \texttt{en\_context} is limited to 32 tokens. Furthermore, we set $\lambda_{\mathrm{cv}} = 0.1$ and $\lambda_{\mathrm{ord}} = 0.2$. The temperature parameter $\tau$ is set to 0.01 and 0.02 for the \texttt{cvccl} and \texttt{oscl} losses, respectively. 

For the Ridge regression meta-model, we consider a set of candidate regularization parameters $\alpha \in \{0.01, 0.1, 1.0, 10.0, 100.0\}$ and select the best value via cross-validation. Finally, We use Root Mean Squared Error (RMSE) as the primary metric for system ranking and report Pearson correlation for completeness.

\section{Ordinal Structure Analysis Details}
\label{sec:ordinal_structure}

To quantitatively assess whether the learned representations preserve the ordinal structure of lexical difficulty, we analyze the relationship between embedding distances and differences in ground-truth difficulty scores. Given a set of instances $\{(x_i, y_i)\}_{i=1}^N$, where $y_i$ denotes the GLMM difficulty score, we first obtain normalized sentence embeddings $\mathbf{h}_i \in \mathbb{R}^d$. For a sampled subset of size $N'$, we compute all pairwise cosine distances:
\[
d_{ij} = 1 - \frac{\mathbf{h}_i \cdot \mathbf{h}_j}{\|\mathbf{h}_i\| \|\mathbf{h}_j\|}
\]
and corresponding absolute difficulty differences:
\[
\Delta_{ij} = |y_i - y_j|.
\]

We then group instance pairs into $K$ bins based on quantiles of $\Delta_{ij}$ and compute the mean embedding distance within each bin. This results in a function $f(k)$ that maps increasing difficulty differences to average embedding distances. A monotonically increasing trend in $f(k)$ indicates that the representation space is structured according to difficulty.

\section{Cross-lingual Alignment Analysis Details}
\label{sec:cross_lingual_details}

To evaluate whether the learned representations capture cross-lingual semantic alignment, we measure the cosine similarity between representations of semantically corresponding inputs across languages. For each instance $x_i$, we construct two views: an L1-based input and its corresponding English-based input. We then obtain normalized embeddings $\mathbf{h}_i^{\text{L1}}$ and $\mathbf{h}_i^{\text{EN}}$ using the encoder.

We define the similarity of aligned pairs as:
\[
s_i^{\text{aligned}} = \mathbf{h}_i^{\text{L1}} \cdot \mathbf{h}_i^{\text{EN}},
\]
and construct a baseline by randomly permuting the English representations to obtain mismatched pairs:
\[
s_i^{\text{random}} = \mathbf{h}_i^{\text{L1}} \cdot \mathbf{h}_{\pi(i)}^{\text{EN}},
\]
where $\pi(i)$ is a random permutation. By comparing the distributions of $s^{\text{aligned}}$ and $s^{\text{random}}$, we assess whether the model places semantically corresponding inputs closer in the embedding space than unrelated ones.

\end{document}